\documentclass{article}

\usepackage{arxiv}

\usepackage[utf8]{inputenc}
\usepackage{bm}
\usepackage{color, soul}
\usepackage{framed}
\usepackage{color}
\usepackage{tikz}
\usepackage{pgflibraryarrows}
\usepackage{pgflibrarysnakes}
\usepackage{booktabs}
\usepackage{hyperref}
\hypersetup{
    colorlinks=true,
    linkcolor=blue,
    filecolor=magenta,      
    urlcolor=cyan,
    pdftitle={Overleaf Example},
    pdfpagemode=FullScreen,
    }
\title{Learning Deep Semantic Model for Code Search using CodeSearchNet Corpus}
\author{Chen Wu\thanks{~~Corresponding author}\\
  Tencent Cloud\\
  \texttt{chewu@tencent.com}\And

  Ming Yan \\
  Alibaba Group\\
  \texttt{ym119608n@alibaba-inc.com}
 }

\begin{document}
\sethlcolor{yellow}
\maketitle

\begin{abstract}
\setlength{\parindent}{0pt} \setlength{\parskip}{1.5ex plus 0.5ex minus 0.2ex} 
Semantic code search is the task of retrieving relevant code snippet given a natural language query. 
Different from typical information retrieval tasks, code search requires to bridge the semantic gap between the programming language and natural language, for better describing intrinsic concepts and semantics.
Recently, deep neural network for code search has been a hot research topic. 
Typical methods for neural code search first represent the code snippet and query text as separate embeddings, and then use vector distance (e.g. dot-product or cosine) to calculate the semantic similarity between them.
There exist many different ways for aggregating the variable length of code or query tokens into a learnable embedding, including bi-encoder, cross-encoder, and poly-encoder. 
The goal of the query encoder and code encoder is to produce embeddings that are close with each other for a related pair of query and the corresponding desired code snippet, in which the choice and design of encoder is very significant. 

In this paper, we propose a novel deep semantic model which makes use of the utilities of not only the multi-modal sources, but also feature extractors such as self-attention, the aggregated vectors, combination of the intermediate representations. We apply the proposed model to tackle the CodeSearchNet challenge \cite{husain2019codesearchnet} about semantic code search.
We align cross-lingual embedding for multi-modality learning with large batches and hard example mining, and combine different learned representations for better enhancing the representation learning.
Our model is trained on CodeSearchNet corpus and evaluated on the held-out data, the final model achieves 0.384 NDCG and won the first place in this benchmark.
Models and code are available at \url{https://github.com/overwindows/SemanticCodeSearch.git}.

\end{abstract}

\section{Introduction}
Recently, We have witnessed increasing interests in natural language to code search. 
More and more people use search engines like Goolge and Bing to search for the code snippet with text query. The ability to retrieve relevant code snippet to a developer's intent is very essential.
Sites such as GeekforGeek and Stack Overflow gain their popularity because they are easy to search for code relevant to user's natural question. 
The goal of code search is to retrieve code snippets from a large code corpus that best match a developer's intent, which is expressed in natural language.
Searching for code is one of the most common tasks for software developers, while search engine results are often frustrating.
Fortunately, recent works from both industry and academia have taken steps towards enabling advanced code search by using deep learning. 
These have been driven by the advent of large datasets, substantial computational capacity, and a number of advances in machine learning models.

The CodeSearchNet challenge is defined on top of of the CodeSearchNet Corpus which encourages academic and industry researchers to study this interesting task and test out new ideas on the semantic code search.
The organizers collect the corpus from publicly available open-source GitHub repositories, using libraries.io to identify all projects which are used by at least one other project, and sort them by “popularity” as indicated by the number of stars and forks. 
Then, they remove all the projects that do not have a license or whose license does not explicitly permit the re-distribution of parts of the project. The detailed descriptions can be found in \cite{husain2019codesearchnet}.

The goal of this challenge is to predict search relevance of code snippets according to query search. In the search challenge, both the “multiple languages” and “single language” scenarios are considered. The final evaluation performance is a mean NDCG on six programming languages (Go, Java, JavaScript, PHP, Python, and Ruby).

We define the challenge as a ranking problem and adopt the learning to rank approach. This paper describes our solution at this competition. 
The remainder of this paper is organized as follows, Section 2 briefly reviews related work. Section 3 describes our model for code search. The result is shown in section 4, followed by the conclusion in section 5.

\section{Related Work}
We formulate code search as a multi-modality task. Our work is based on two recent extensions to the latent semantic models for IR. The first is the exploration of deep learning methods for semantic modeling. The second is the introduction of multi-modality latent interaction.

\begin{table}
  \caption{Human Language Vs Programming Language}
  \label{tab:freq}
  \centering
  \begin{tabular}{cccc}
  \toprule
    \textbf{Human}& \textbf{Machine}\\
  \midrule
    Mutual understanding  & Machine task completion \\
    Closed vocabulary & Open vocabulary \\
    Imprecise grammar & Precise grammar \\
    Ambiguity expected & Ambiguity is unacceptable \\
  \bottomrule
\end{tabular}
\end{table}

\subsection{Text Retrieval-based Code Search}
A conventional Lucene-based search tool could be used for code search. 
The main problem is that the query expressed with natural language is quite different from the source code snippets that contain the intended results. 
An effective code search engine should be able to understand the semantic meanings of natural language queries and source code in order to improve the accuracy of code search.

\subsection{Deep Learning-based Code Search}
Joint Embedding (also known as multi-model embedding) 
is a technique to jointly embed heterogeneous data into a unified vector space so that semantically similar concepts across different modalities occupy nearby regions of the space.


Neural code search model learns embeddings such that a query description and its corresponding code snippet are both mapped to a similar point in the same shared embedding space. Then given a natural language query, it can embed the query in the  vector space and search for nearby code snippets. 
Compared to text retrieval-powered code search tools, the recently proposed deep learning-based methods give excellent results.




\subsection{Neural Architectures}
We mainly consider three different neural model architectures listed as below: bi-encoder, cross-encoder, poly-encoder.

A fundamental trade-off between effectiveness and efficiency needs to be balanced when designing an online search engine.
Effectiveness comes from sophisticated functions such as cross attention, while efficiency is obtained from improvements in preliminary retrieval components such as representation pre-computing and embedding indexing. Given the complexity of the real-world code search scenario, it is difficult to jointly optimize both in an end-to-end system. To address this problem, we develop a novel deep learning model based on bi-encoder.




\section{Deep Semantic Code Search}
\subsection{Task Definition}
Given a text query in natural language as input, the task aims to find the most semantically related code from a collection of candidate codes. 
Based on this, given a query or question, the machine needs to first read and understand the query, and then finds the code snippet to answer the query according to the relevance score. 
The query is described as a sequence of word tokens $q=\{w_{t}^{q}\}_{t=1}^{n}$ and the code snippet is described as $c=\{w_{t}^{c}\}_{t=1}^{m}$ where n is the number of tokens in the query and m is the number of words in the candidate code snippet. 
The relevance score $r$ is designed as the semantic similarity between a query $q$ and a code snippet $c$. 
The object function for semantic code search is to learn a function $f(q,c)={argmax}_{c\in CSN} P(c|q)$, where $CSN$ is the total CodeSearchNet corpus. 
The training data is a set of the relevant query and code tuples $\left \langle q,c \right \rangle$ from CodeSearchNet corpus.

\subsection{Deep Match Framework}
\begin{figure}
    \centering
    \includegraphics[width=1.0\textwidth]{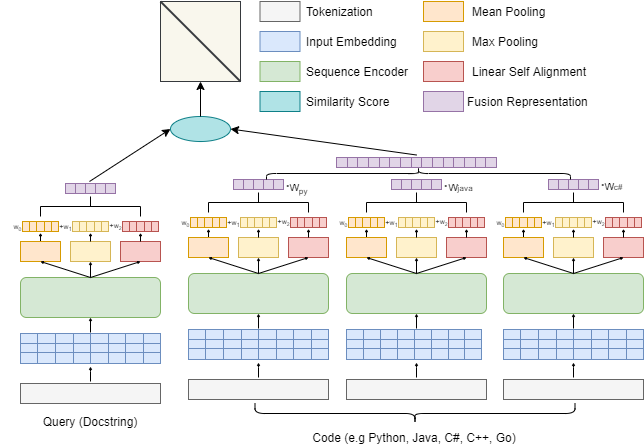}
    \caption{Framework}
    \label{fig:arch}
\end{figure}

Next We will describe our framework from the bottom up. 
The proposed model consists of three main components: a code embedding network to embed source code snippets as vectors, a description embedding network to embed natural language descriptions as vectors, and a cosine similarity module to measure the degree of similarity between code and descriptions. 
The overall framework is shown in Figure \ref{fig:arch}.

\textbf{Encoder Layer} 
could be seen as a language model which utilizes contextual cues from surrounding words.
Encoder layer converts the query and code snippet from tokens to the semantic representations respectively.
There are many popular architectures such as CBoW, LSTM, CNN and Transformer Encoder. We take CBoW as our encoder architecture in terms of efficiency.

The query embedding network encode the natural language description from the user query in online inference, while in training stage the method comment will be used as ``virtual'' query.
Both the descriptions of query and code snippet (a method) are first turned into a set of tokens, where We leverage BPE to do the tokenization.
In CBoW, the tokens have no strict order and they are embedded using a conventional multi-layer perception network, after which are further aggregated using a pooling method. 

\textbf{Match Layer} 
employs a match function to compute the relevance between the query and code representation on method level.

\textbf{Output Layer} 
uses pairwise or list-wise ranking function to rank the code snippet based on the relevance score to the given query. 

The main contribution of this work lies in the novel encoder layer, where a hybrid/fusion pooling strategy is used to make use of different aspects of the representations, in order to capture the semantic representations of natural language and programming language. 
A linear self-alignment function is introduced to better obtain the contextual representations. 
The detailed description of the model is provided as follows.

\subsection{Aligning Contextual Embedding}
Embedding alignment was originally studied for word vectors with the goals of enabling cross-lingual transfer, where the embeddings for two languages are in alignment with word translations. 
Multi-modal learning is a challenging task as it has different types of inputs such as code snippet, natural language text.
In our work, code token vectors are learnt independently for each language, and we learn a linear map $W_{lang}$ for each language and use it to project the code token vectors to a same semantic space, as defined in Equation \ref{equ:1}.

\begin{equation}
    \label{equ:1}
    e^{'}=\mathbf{W_{lang}}\mathbf{e}
\end{equation}

\subsection{Self Attention Pooling}
Pooling is an important component of a wide variety of sentence representation and embedding models.
Different from the widely used max-pooling and mean-pooling, we explore inner/self-sentence attention mechanism for sentence embedding. We follow the encoding method used in \cite{wang2018multi} and adopt a linear transformation to encode the query and code snippet representation to a single vector, respectively.

First, the encoder network is applied on the query/code inputs. Then we aggregate the resulting hidden outputs into one single vector as Equation \ref{equ:2}, with a linear self-alignment as Equation \ref{equ:3}:

\begin{equation}
    \label{equ:2}
    \gamma={softmax}\left(\mathbf{w}_{\mathrm{e}}^{\top} \cdot \mathrm{E}^{\prime}\right)
\end{equation}

\begin{equation}
    \label{equ:3}
    \mathbf{e^{\prime \prime}}=\sum_{\mathrm{j}} \gamma_{\mathrm{j}} \cdot \mathrm{E}_{: \mathrm{j}}^{\prime}, \forall \mathrm{j} \in[1, \ldots, \mathrm{m}]
\end{equation}
where $\mathbf{w}_{\mathrm{e}}$ is a weight vector to learn, we self-align the refined query/code representations into a single vector according to the learnt self-attention weight, which can be further used to compute the matching with the code/query representations.

\subsection{Fusion Representations}
Inspired by ELMo language model \cite{peters2018deep}, we propose a hybrid method to fuse all aggregated vectors in encoder part into a single vector, and then compute language-specific weights to generate a fusion representation:

\begin{equation}
\label{equ:4}
\mathbf{FUSED}_{k}^{t a s k}=E\left(R_{k} ; \Theta^{t a s k}\right)=\gamma^{t a s k} \sum_{j=0}^{L} s_{j}^{t a s k} \mathbf{h}_{k, j}^{L M}
\end{equation}

In Equation \ref{equ:4}, $s^{task}_j$ are softmax-normalized weights and the scalar parameter $\Theta^{task}$ allows the language encoder to scale the entire fused vector.

\subsection{Tokenization}
According to CodeSearchNet \cite{husain2019codesearchnet}, We tokenize all Go, Java, JavaScript, Python, PHP and Ruby functions (or methods) using TreeSitter and BPE, where available, their respective documentation text using a heuristic regular expression.

We use TreeSitter parser to parse the code snippet which help to identify the function name, variable name and keywords etc. Based on the tree-sitter results, we further apply BPE tokenization to reduce the vocabulary size.

\subsection{Learning the Model}
During training, we use a similarity measure (e.g. cosine) in the loss function to encourage the joint mapping of query and code.

Training instances are given as triples (C, D+, D-), where C is a code snippet, D+ is the correct (actual) description of C, and D- is an incorrect description, where D- is randomly chosen from the pool of all descriptions. The loss function seeks to maximize the cosine similarity between C and D+, and make the distance between C and D- as large as possible.

The training corpus is based on the corpus on CodeSearchNet.

\section{Indexing and querying}
To index a codebase, we embed all code snippets in the codebase into vectors via offline processing. Then for online searching, we embed the user query, and estimate the cosine similarities between the query embedding and pre-computed code snippet embeddings. The top K closest code snippets are returned as the query results.

\section{Experiments}

\section{DataSet Overview}
The CodeSearchNet dataset consists of 2 million comment-code pairs from open source libraries. Specifically, a comment is a top-level function or method comment (e.g. docstrings in Python), and code is an entire function or method. Currently, the dataset contains Python, Javascript, Ruby, Go, Java, and PHP code. Throughout this paper, we refer to the terms docstring/fucntion, comments and query interchangeably. We partition the data into train, validation, and test splits such that code from the same repository can only exist in one partition. Currently this is the only dataset on which we train our model. Summary statistics about this dataset can be found in Table \ref{tab:dataset}.

\begin{table}
  \caption{CodeSearchNet}
  \label{tab:dataset}
  \centering
  \begin{tabular}{lllllll}
  \toprule
    \textbf{Partition} & \textbf{Go} & \textbf{Java} & \textbf{JavaScript} & \textbf{Php} & \textbf{Python} & \textbf{Ruby}\\
  \midrule
    Train & 317832 & 454451 & 123889 & 523712 &  412178 & 48791 \\
    Test & 14291 & 26909 & 6483 & 28391 & 22176 & 2279  \\
    Valid & 14242 & 15328 & 8253 & 26015 & 23107 & 2209\\
  \bottomrule
\end{tabular}
\end{table}

\subsection{Evaluation Methodology}
The metric we use for evaluation is Normalized Discounted Cumulative Gain (NDCG).
All code is indexed (including methods without any comments), resulting in just over-indexed methods.
To qualify, the questions must be about a concrete programming task and include a code snippet in the accepted answer. 

\subsection{Results}
The performance of the proposed model with ablation study is shown in table \ref{tab:freq}. 

\begin{table}
  \caption{Benchmark}
  \label{tab:freq}
  \centering
  \begin{tabular}{llllllll}
  \toprule
    \textbf{Model} & \textbf{NDCG} & \textbf{Go} & \textbf{Java} & \textbf{JavaScript} & \textbf{Php} & \textbf{Python} & \textbf{Ruby}\\
  \midrule
    NBoW+Weighted & 0.3841 & 0.3253 & 0.4248 & 0.3611 &  0.3399 & 0.4717 & 0.3816 \\
    NBoW+Self-Attn & 0.3732 & 0.3415 & 0.4075 & 0.3455 & 0.3235 & 0.4551 & 0.366 \\
    NBoW Baseline & 0.340 & 0.278 & 0.355 & 0.311 & 0.291 & 0.448 & 0.360   \\
  \bottomrule
\end{tabular}
\end{table}

\section{Conclusion}
In this paper, we propose a novel deep semantic model for code search. Aligning cross-lingual embedding for multi-modality learning and combining different learned representations with novel self-attention pooling method are introduced to better understand the query and code snippets. Experimental results demonstrate that the model can obtain new state-of-the-art results in the popular CodeSearchNet benchmark.

\section{Acknowledgement}
Thank the organizers of CodeSearchNet and Github for generously providing the opportunity and data resources for training and testing our solutions.

\bibliographystyle{plain}
\bibliography{references}   
\end{document}